\documentclass[runningheads]{llncs}
\usepackage{graphicx}
\usepackage{comment}
\usepackage{booktabs}       
\usepackage{multirow}

\usepackage{subfigure}

\usepackage{bbding}

\usepackage{amsmath}
\usepackage{amsfonts}
\usepackage{amssymb}

\usepackage{wrapfig}

\newcommand{\RN}[1]{%
	\textup{\lowercase\expandafter{\it \romannumeral#1}}%
}

\usepackage{xcolor}
\definecolor{mygreen}{HTML}{3cb44b}
\definecolor{skyblue}{HTML}{beffff}
\definecolor{lightgreen}{HTML}{90ee90}

\newcommand{\ie}[0]{\emph{i.e., }}
\newcommand{\ea}[0]{\emph{et al. }}
\newcommand{\eg}[0]{\emph{e.g., }}

\newcommand{\etc}[0]{\emph{etc.}}

\newcommand{\beq}{\vspace{0mm}\begin{equation}}
\newcommand{\eeq}{\vspace{0mm}\end{equation}}
\newcommand{\beqs}{\vspace{0mm}\begin{eqnarray}}
\newcommand{\eeqs}{\vspace{0mm}\end{eqnarray}}
\newcommand{\barr}{\begin{array}}
\newcommand{\earr}{\end{array}}

\newcommand{\Imat}{{\bf I}}

\newcommand{\Wmat}[0]{{{\bf W}}}

\newcommand{\hv}[0]{{\boldsymbol{h}}}

\newcommand{\qv}[0]{{\boldsymbol{q}}}

\newcommand{\vv}{\boldsymbol{v}}
\newcommand{\wv}{\boldsymbol{w}}
\newcommand{\xv}{\boldsymbol{x}}

\newcommand{\thetav}{\boldsymbol{\theta}}

\newcommand{\R}{\mathbb{R}}
\newcommand{\E}{\mathbb{E}}

\usepackage{caption}

\newcommand{\Lcal}{\mathcal{L}}

\newcommand{\Dcal}{\mathcal{D}}

\definecolor{Gray}{gray}{0.93}
\definecolor{Graylight}{gray}{0.95}
\definecolor{Grayheavy}{gray}{0.90}

\usepackage{colortbl}
\definecolor{Gray}{gray}{0.93}

\newcommand{\short}{\textsc{Oscar}}
\newcommand{\shortb}{\textsc{Oscar}$_{\text{B}}$}
\newcommand{\shortl}{\textsc{Oscar}$_{\text{L}}$}
\newcommand{\longname}{\textbf{O}bject-\textbf{S}emanti\textbf{c}s \textbf{A}ligned P\textbf{r}e-training}

\def\secvspace{{\vspace{-4mm}}}
\def\suppvspace{{\vspace{-3mm}}}

\begin{document}
\pagestyle{headings}
\mainmatter
\def\ECCVSubNumber{7133}  

\title{\short: Object-Semantics Aligned Pre-training for Vision-Language Tasks} 

%

\titlerunning{\short{}: Object-Semantics Aligned Pre-training for Vision-Language Tasks}
%
\author{
Xiujun Li\textsuperscript{$\heartsuit\spadesuit$}
\and Xi Yin\textsuperscript{$\heartsuit$}
\and Chunyuan Li\textsuperscript{$\heartsuit$} 
\and Pengchuan Zhang\textsuperscript{$\heartsuit$} 
\and Xiaowei Hu\textsuperscript{$\heartsuit$} 
\and \\
Lei Zhang\textsuperscript{$\heartsuit$} 
\and Lijuan Wang\textsuperscript{$\heartsuit$} 
\and Houdong Hu\textsuperscript{$\heartsuit$} 
\and Li Dong\textsuperscript{$\heartsuit$}
\and Furu Wei\textsuperscript{$\heartsuit$} 
\and \\
Yejin Choi\textsuperscript{$\spadesuit$} 
\and Jianfeng Gao\textsuperscript{$\heartsuit$}
}
\authorrunning{X. Li, X. Yin, C. Li et al.}
%
\institute{
 \textsuperscript{$\heartsuit$}Microsoft Corporation
\hspace{10mm}
 \textsuperscript{$\spadesuit$}University of  Washington
}
\maketitle

\begin{abstract}
Large-scale pre-training methods of learning cross-modal representations on image-text pairs are becoming popular for vision-language tasks. While existing methods simply concatenate image region features and text features as input to the model to be pre-trained and use self-attention to learn image-text semantic alignments in a brute force manner, in this paper, we propose a new learning method~\short{}\footnote{\longname{}}, which uses object tags detected in images as \emph{anchor points} to significantly ease the learning of alignments. Our method is motivated by the observation that the salient objects in an image can be accurately detected, and are often mentioned in the paired text. 
%
We pre-train an \short{} model on the public corpus of $6.5$ million text-image pairs, and fine-tune it on downstream tasks, creating new state-of-the-arts on six well-established vision-language understanding and generation tasks.\footnote{The code and pre-trained models are released: \url{https://github.com/microsoft/Oscar}}

\end{abstract}

\section{Introduction}
Learning cross-modal representations is fundamental to a wide range of vision-language (V+L) tasks, such as visual question answering, image-text retrieval, image captioning. Recent studies~\cite{lu2019vilbert,tan2019lxmert,chen2019uniter,su2019vl,li2019visualbert,li2019unicoder,zhou2019unified} on vision-language pre-training (VLP) have shown that it can effectively learn generic representations from massive image-text pairs, and that fine-tuning VLP models on task-specific data achieves state-of-the-art (SoTA) results on well-established V+L tasks.

These VLP models are based on multi-layer Transformers~\cite{vaswani2017attention}.
To pre-train such models, existing methods simply concatenate image region features and text features as input and resort to the self-attention mechanism to learn semantic alignments between image regions and text in a brute force manner.
However, the lack of explicit alignment information between the image regions and text poses alignment modeling a weakly-supervised learning task. In addition, visual regions are often over-sampled \cite{anderson2018bottom}, noisy and ambiguous, which makes the task even more challenging. 

\begin{figure}[t!]
\centering
{\includegraphics[width=0.90\textwidth]{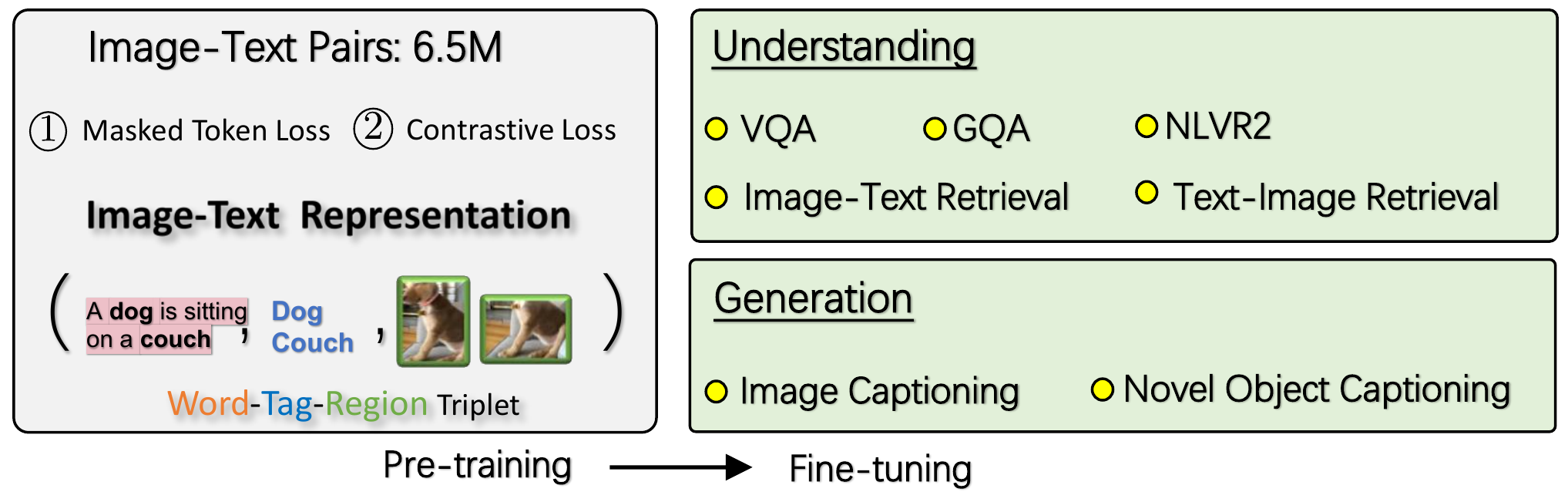}}
\vspace{-3mm}
\caption{\small \short{} pipeline. The model takes a triple as input, is pre-trained with two losses (a masked token loss over words \& tags, and a contrastive loss between tags and others), and fine-tuned for 5 understanding and 2 generation tasks (detailed in Sec. 4). }
\label{fig:pretrain_finetune}
\vspace{-5mm}
\end{figure}

In this study, we show that the learning of cross-modal representations can be significantly improved by introducing object tags detected in images as \emph{anchor points} to ease the learning of semantic alignments between images and texts. 
We propose a new VLP method \short, where we define the training samples as triples, each consisting of a word sequence, a set of object tags, and a set of image region features. Our method is motivated by the observation that the salient objects in an image can be accurately detected by modern object detectors~\cite{ren2015faster}, and that these objects are often mentioned in the paired text.
For example, on the MS COCO dataset~\cite{lin2014microsoft}, the percentages that an image and its paired text share at least $1$, $2$, $3$ objects are $49.7\%$, $22.2\%$, $12.9\%$, respectively. Our \short{} model is pre-trained on a large-scale V+L dataset composed of $6.5$ million pairs, and is fine-tuned and evaluated on seven V+L understanding and generation tasks. The overall setting is illustrated in Fig~\ref{fig:pretrain_finetune}.

Although the use of anchor points for alignment modeling has been explored in natural language processing~\eg~\cite{brown1991aligning}, to the best of our knowledge, this work is the first that explores the idea for VLP. There have been previous works that use object or image tags in V+L tasks for the sake of enhancing the feature representation of image regions, rather than for learning image-text alignments. For example, Zhou \ea~\cite{zhou2019unified} uses the object prediction probability as a soft label and concatenate it with its corresponding region features. Wu \ea~\cite{wu2016value} and You \ea~\cite{you2016image} introduce image-level labels or attributes to improve image-level visual representations.

The main contributions of this work can be summarized as follows:
$(\RN{1})$ We introduce \short{}, a powerful VLP method to learn generic image-text representations for V+L understanding and generation tasks.
$(\RN{2})$ We have developed an \short{} model that achieves new SoTA on multiple V+L benchmarks, outperforming existing approaches by a significant margin; 
$(\RN{3})$ We present extensive experiments and analysis to provide insights on the effectiveness of using object tags as anchor points for cross-modal representation learning and downstream tasks.






\vspace{-4mm}
\section{Background}
\vspace{-2mm}

\begin{figure*}[t!]
\centering
\begin{tabular}{c c c}
	\hspace{-2mm}
	\includegraphics[height=2.5cm]{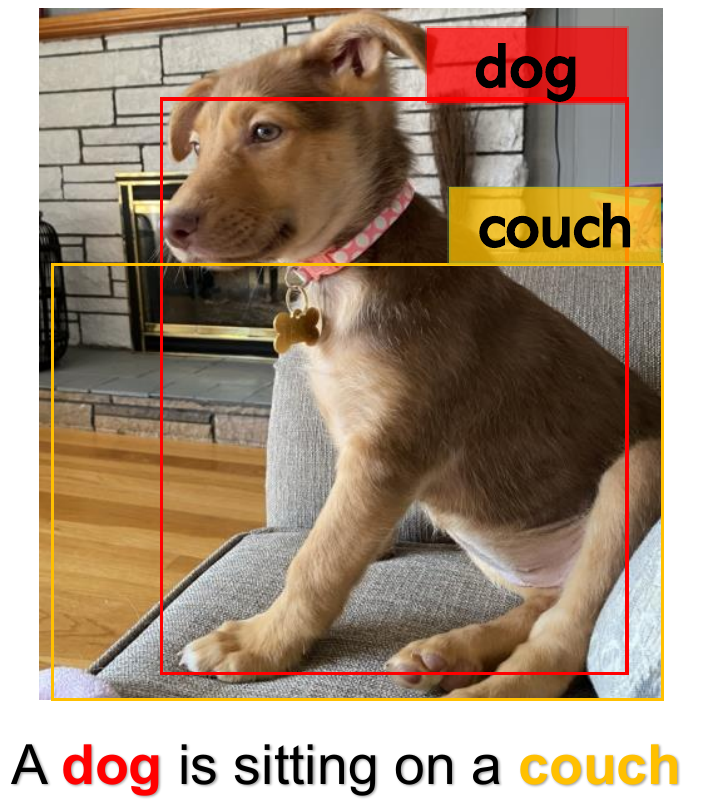} & 
	 \hspace{-0mm}
	\includegraphics[height=2.5cm]{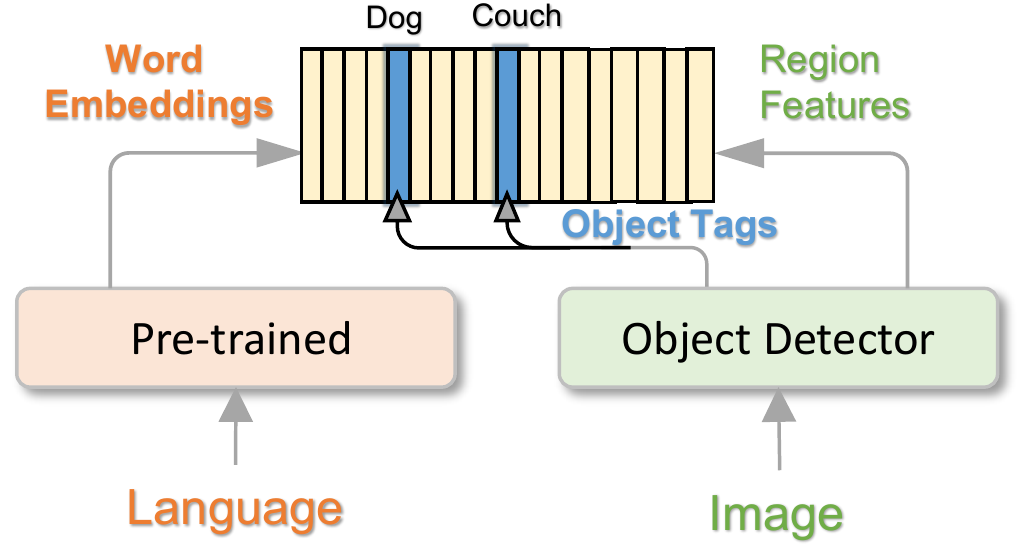} &
	\hspace{2mm}  
	\includegraphics[height=2.5cm]{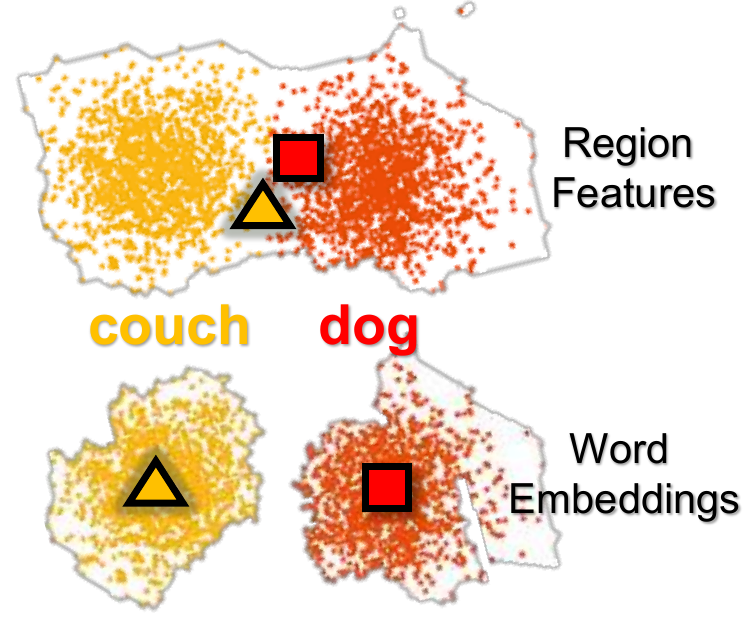}
	\\
	(a) Image-text pair \vspace{2mm} & 
	\hspace{-2mm}  
	(b) Objects as anchor points \hspace{-0mm} &
	(c) Semantics spaces
	\\ 
\end{tabular}
\vspace{-2mm}
\caption{Illustration on the process that \short{} represents an image-text pair into semantic space via dictionary look up. (a) An example of input image-text pair (b) The object tags are used as anchor points to align image regions with word embeddings of pre-trained language models. (c) The word semantic space is more representative than image region features. In this example, $\mathtt{dog}$ and $\mathtt{couch}$ are similar in the visual feature space due to the overlap regions, but distinctive in the word embedding space.}
\label{fig:process}
\vspace{-2mm}
\end{figure*}

The training data for many V+L tasks consists of image-text pairs, as shown in Fig.~\ref{fig:process}(a). We denote a dataset of size $N$ by $\Dcal = \{ (\Imat_i, \wv_i) \}_{i=1}^N$, with image $\Imat$ and text sequence $\wv$. The goal of pre-training is to learn cross-modal representations of image-text pairs in a self-supervised manner, which can be adapted to serve various down-stream tasks via fine-tuning. 

VLP typically employs multi-layer self-attention Transformers~\cite{vaswani2017attention} to learn cross-modal {\it contextualized} representations, based on the {\it singular} embedding of each modality. 
Hence, the success of VLP fundamentally relies on the quality of the input singular embeddings. 
Existing VLP methods take visual region features $\vv = \{ v_1, \cdots, v_K\}$ of an image and word embeddings $\wv = \{ w_1, \cdots, w_T \}$ of its paired text as input, and relies on the self-attention mechanism to learn image-text alignments and produce cross-modal contextual representations.


Though intuitive and effective, 
existing VLP methods suffer from two issues: 
$(\RN{1})$ {\it Ambiguity}. The visual region features are usually extracted from over-sampled regions~\cite{anderson2018bottom} via Faster R-CNN object detectors~\cite{ren2015faster}, which inevitably results in overlaps among image regions at different positions. 
This renders ambiguities for the extracted visual embeddings. For example, in Fig.~\ref{fig:process}(a) the region features for $\mathtt{dog}$ and $\mathtt{couch}$ are not easily distinguishable, as their regions heavily overlap.
%
$(\RN{2})$ {\it Lack of grounding}. VLP is naturally a weakly-supervised learning problem because there is no explicitly labeled alignments between regions or objects in an image and words or phrases in text. 
However, we can see that salient objects such as $\mathtt{dog}$ and $\mathtt{couch}$ are presented in both image and its paired text as in Fig.~\ref{fig:process}(a), and can be used as anchor points for learning semantic alignments between image regions and textual units as in Fig.~\ref{fig:process}(b). In this paper we propose a new VLP method that utilizes these anchor points to address the aforementioned issues.


\vspace{-3mm}
\section{\short{} Pre-training}
\begin{figure}[t!]
\centering
\includegraphics[width=0.9\columnwidth]{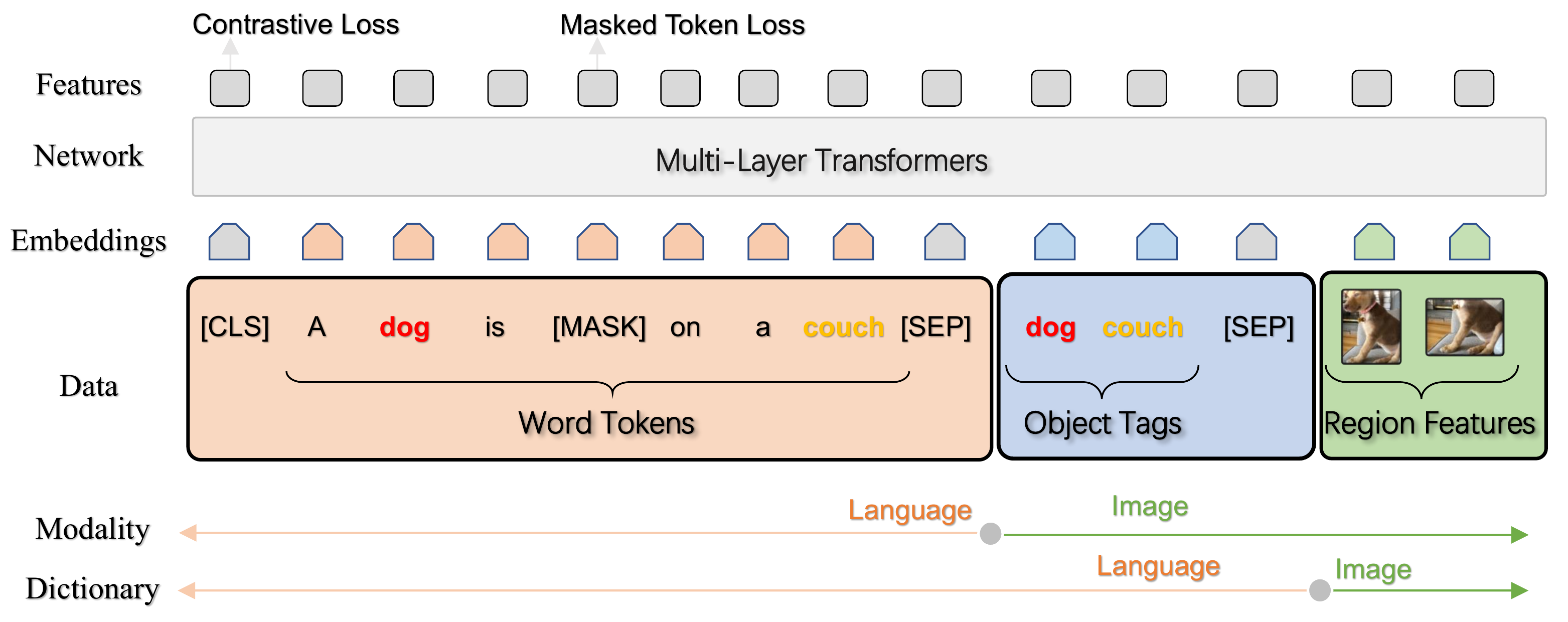}
\caption{Illustration of \short{}. We represent the image-text pair as a triple [\colorbox{red!10}{word tokens},  \colorbox{blue!10}{object tags}, \colorbox{mygreen!10}{region features}], where the object tags (\eg ``\texttt{dog}'' or ``\texttt{couch}'') are proposed to align the cross-domain semantics; when removed, \short{} reduces to previous VLP methods. The input triple can be understood from two perspectives: a {\it modality} view and a {\it dictionary} view.}
\label{fig:oscar}
\end{figure}

\vspace{-3mm}
Humans perceive the world through many channels. Even though any individual channel might be incomplete or noisy, important factors are still perceivable since they tend to be shared among multiple channels (\eg $\mathtt{dog}$ can be described visually and verbally, as in Fig.~\ref{fig:process}). With this motivation, we propose a new VLP method \short{} to learn representations that capture channel-invariant (or modality-invariant) factors at the semantic level.
Oscar differs from existing VLP in the way that the input image-text pairs are represented and the pre-training objective, as outlined in Fig.~\ref{fig:oscar}.

\subsubsection{Input}

\short{} represents each input image-text pair as a Word-Tag-Image triple $(\wv, \qv, \vv)$, where $\wv$ is the sequence of word embeddings of the text, $\qv$ is the word embedding sequence of the object tags (in text) detected from the image, and $\vv$ is the set of region vectors of the image.

Existing VLP methods represent each input pair as $(\wv, \vv)$. \short{} introduces $\qv$ as anchor points to ease the learning of image-text alignment. This is motivated by the observation that in training data, important objects in an image are often also \emph{presented} in the image-paired text, using either the same words as object tags or different but semantically similar or related words. Since the alignments between $\qv$ and $\wv$, both in text, are relatively easy to identified by using pre-trained BERT models~\cite{devlin2019bert}, which are used as initialization for VLP in \short, the image regions from which the object tags are detected are likely to have higher attention weights than other regions, when queried by the semantically related words in the text. This alignment learning process is conceptually illustrated in Fig.~\ref{fig:process}(b). The process can also be interpreted as learning to ground the image objects, which might be ambiguously represented in the vision space such as $\mathtt{dog}$ and $\mathtt{couch}$ in Fig.~\ref{fig:process}(a), in distinctive entities represented in the language space, as illustrated in Fig.~\ref{fig:process}(c).
 
Specifically, $\vv$ and $\qv$ are generated as follows.
Given an image 
with $K$ regions of objects (normally over-sampled and noisy), Faster R-CNN~\cite{ren2015faster} is used to extract the visual semantics of each region as $(v^{\prime}, z)$, where region feature $v^{\prime} \in \R^P$ is a $P$-dimensional vector (\ie $P=2048$), and region position $z$ a $R$-dimensional vector (\ie $R=4$ or $6$)\footnote{It includes coordinates of top-left \& bottom-right corners, and/or height \& width.}. We concatenate $v^{\prime}$ and $z$ to form a position-sensitive region feature vector, which is further transformed into $v$ using a linear projection to ensure that it has the same vector dimension as that of word embeddings.
Meanwhile, the same Faster R-CNN is used to detect a set of high precision object tags. 
$\qv$ is the sequence of word embeddings of the object tags.

\subsubsection{Pre-Training Objective}

The \short{} input can be viewed from two different perspectives as

\begin{align} 
\xv 
\triangleq [\underbrace{~\wv_{~}}_{\text{\textcolor{red!50}{language}} }, \underbrace{~\qv, \vv~}_{\text{\textcolor{mygreen}{image}}}]
= [ \underbrace{~\wv, \qv}_{\text{\textcolor{red!50}{language}}},~ \underbrace{~\vv_{~_{~}}}_{\text{\textcolor{mygreen}{image}}}]  
\triangleq \xv^{\prime}
\label{eq_two_views}
\end{align}
where $\xv$ is the {\it modality} view to distinguish the representations between a text and an image; while $\xv^{\prime}$ is the {\it dictionary} view\footnote{A semantic space can be viewed a vector space defined by a dictionary, which maps an input to a vector representation in the semantic space. For example, BERT can be viewed as a dictionary that defines a linguistic semantic space. BERT maps an input word or word sequence into a feature vector in the semantic space.}
to distinguish the two different semantic spaces, in which the input is represented. The two-view perspective allows us to design a novel pre-training objective.

\paragraph{A Dictionary View: Masked Token Loss.} 
The use of different dictionaries determines the semantic spaces utilized to represent different sub-sequences. Specifically,
the object tags and word tokens share the same linguistic semantic space, while the image region features lie in the visual semantic space. 
We define the {\it discrete token sequence} as $\hv \triangleq [\wv, \qv]$, and apply the Masked Token Loss (MTL) for pre-training. At each iteration, we randomly mask each input token in $\hv$ with probability $15\%$, and replace the masked one $h_i$ with a special token $\mathtt{[MASK]}$. The goal of training is to predict these masked tokens based on their surrounding tokens $\hv_{\backslash i}$ and all image features $\vv$ by minimizing the negative log-likelihood:
\begin{align}
\hspace{-0mm}
\Lcal_{\text{MTL}} = -\E_{  (\vv, \hv) \sim \Dcal } \log p( h_i | \hv_{\backslash i}, \vv )
\label{eq_attend_mlm}
\end{align}
This is similar to masked language model used by BERT. The masked word or tag needs to be recovered from its surroundings, with additional image information attended to help ground the learned word embeddings in the vision context.

\paragraph{A Modality View: Contrastive Loss.} 
For each input triple, we group $\hv^{\prime}\triangleq[\qv, \vv]$ to represent the image modality, and consider $\wv$ as the language modality. We then sample a set of ``polluted'' image representations by replacing $\qv$ with probability 50\% with a different tag sequence randomly sampled from the dataset $\Dcal$. Since the encoder output on the special token $\mathtt{[CLS]}$ is the fused vision-language representation of $(\hv^{\prime}, \wv)$, we apply a fully-connected (FC) layer on the top of it as a binary classifier $f(.)$ to predict whether the pair contains the original image representation ($y=1$) or any polluted ones ($y=0$). 
The contrastive loss is defined as

%
\begin{align}
\Lcal_{\text{C}} = -\E_{  (\hv^{\prime}, \wv ) \sim \Dcal } \log p( y | f(\hv^{\prime}, \wv) ).
\label{eq_action_prediction}
\end{align}
During the cross-modal pre-training, we utilize object tags as the proxy of images to adjust the word embedding space of BERT, where a text is similar to its paired image (or more specifically, the object tags detected from the image), and dissimilar to the polluted ones. 

The full pre-training objective of \short{} is: 
\begin{align}
\Lcal_{\text{Pre-training}} = \Lcal_{\text{MTL}}  + \Lcal_{\text{C}}. 
\label{eq_pre_training}
\end{align}

\paragraph{Discussion.} Although other loss function designs can be considered as pre-training objectives, we perform experiments with these two losses for two reasons: 
$(\RN{1})$  
Each loss provides a representative learning signal from its own perspective. We deliberately keep a clear and simple form for the joint loss to study the effectiveness of the proposed dictionary and modality views, respectively.
$(\RN{2})$
Though the overall loss is much simpler than those of existing VLP methods, it yields superior performance in our experiments.

\subsubsection{Pre-training Corpus}
We have built the pre-training corpus based on the existing V+L datasets, including COCO~\cite{lin2014microsoft}, Conceptual Captions (CC)~\cite{sharma2018conceptual}, SBU captions~\cite{ordonez2011im2text}, flicker30k~\cite{young2014image}, GQA~\cite{hudson2019gqa} \etc. In total, the unique image set is 4.1 million, and the corpus consists of 6.5 million text-tag-image triples. The detail is in Appendix.

%
%

\subsubsection{Implementation Details} 
We pre-train two model variants, denoted as \shortb{} and \shortl{}, initialized with parameters $\thetav_{\text{BERT}}$ of BERT base ($H=768$) and large ($H=1024$), respectively, where $H$ is the hidden size. 
To ensure that the image region features have the same input embedding size as BERT, we transform the position-sensitive region features using a linear projection via matrix $\Wmat$. The trainable parameters are $\thetav=\{\thetav_{\text{BERT}}, \Wmat\}$. The AdamW Optimizer is used. \shortb{} is trained for at least $1.0$M steps, with learning rate $5e^{-5}$ and batch size $768$. \shortl{} is trained for at least $900$k steps, with learning rate $1e^{-5}$ and batch size $512$. The sequence length of discrete tokens $\hv$ and region features $\vv$ are $35$ and $50$, respectively.


\section{Adapting to V+L Tasks} 
We adapt the pre-trained models to seven downstream V+L tasks, including five understanding tasks and two generation tasks. Each task poses different challenges for adaptation. 
We introduce the tasks and our fine-tuning strategy in this section, and leave the detailed description of datasets and evaluation metrics to Appendix.

\secvspace
\subsubsection{Image-Text Retrieval} heavily relies on the joint representations. There are two sub-tasks: {\it image retrieval} and {\it text retrieval}, depending on which modality is used as the retrieved target.
%
%
During training, we formulate it as a binary classification problem. Given an aligned image-text pair, we randomly select a different image or a different caption to form an unaligned pair. The final representation of $\mathtt{[CLS]}$ is used as the input to the classifier to predict whether the given pair is aligned or not. We did not use ranking losses~\cite{karpathy2015deep,lee2018stacked}, as we found that the binary classification loss works better, similarly as reported in~\cite{qi2020imagebert}. In the testing stage, the probability score is used to rank 
the given image-text pairs of a query. Following~\cite{li2019unicoder}, we report the top-$K$ retrieval results on both the $1$K and $5$K COCO test sets.

\secvspace
\subsubsection{Image Captioning}
\label{sec:image_cap} requires the model to generate a natural language description of the content of an image. To enable sentence generation, we fine-tune \short{} using the seq2seq objective. The input samples are processed to triples consisting of image region features, captions, and object tags, in the same way as that during the pre-training. We randomly mask out $15\%$ of the caption tokens and use the corresponding output representations to perform classification to predict the token ids. Similar to VLP~\cite{zhou2019unified}, the self-attention mask is constrained such that a caption token can only attend to the tokens before its position to simulate a uni-directional generation process.  
Note that all caption tokens will have full attentions to image regions and object tags but not the other way around. 

During inference, we first encode the image regions, object tags, and a special token $\mathtt{[CLS]}$ as input. Then the model starts the generation by feeding in a $\mathtt{[MASK]}$ token and sampling a token from the vocabulary based on the likelihood output. Next, the $\mathtt{[MASK]}$ token in the previous input sequence is replaced with the sampled token and a new $\mathtt{[MASK]}$ is appended for the next word prediction. The generation process terminates when the model outputs the $\mathtt{[STOP]}$ token. We use beam search (\ie beam size = 5)~\cite{anderson2018bottom} in our experiments and report our results on the COCO image captioning dataset.

\secvspace
\subsubsection{Novel Object Captioning (NoCaps)} ~\cite{agrawal2019nocaps} extends the image captioning task, and provides a benchmark with images from the Open Images dataset~\cite{kuznetsova2018open} to test models' capability of describing novel objects which are not seen in the training corpus. Following the restriction guideline of NoCaps, we use the predicted Visual Genome and Open Images labels to form tag sequences, and train \short{} on COCO without the initialization of pre-training. 

\secvspace
\subsubsection{VQA}\label{sec:vqa}~\cite{goyal2017making} requires the model to answer natural language questions based on an image. Given an image and a question, the task is to select the correct answer from a multi-choice list. Here we conduct experiments on the widely-used VQA v2.0 dataset~\cite{goyal2017making}, which is built based on the MSCOCO~\cite{lin2014microsoft} image corpus. The dataset is split into training (83k images and 444k questions), validation (41k images and 214k questions), and test (81k images and 448k questions) sets. Following~\cite{anderson2018bottom}, for each question, the model picks the corresponding answer from a shared set consisting of 3,129 answers.

When fine-tuning on the VQA task, we construct one input sequence, which contains the concatenation of a given question, object tags and region features, and then the $\mathtt{[CLS]}$ output from \short{} is fed to a task-specific linear classifier for answer prediction.
We treat VQA as a multi-label classification problem~\cite{anderson2018bottom} – assigning a soft target score to each answer based on its relevancy to the human answer responses, and then we fine-tune the model by minimizing the cross-entropy loss computed using the predicted scores and the soft target scores. At inference, we simply use a Softmax function for prediction.

\secvspace
\subsubsection{GQA}~\cite{hudson2019gqa} is similar to VQA, except that GQA tests the reasoning capability of the model to answer a question. We conduct experiments on the public GQA dataset~\cite{hudson2019gqa}. 
For each question, the model chooses an answer from a shared set of $1,852$ candidate answers. We develop two fine-tuned models using \short$_B$. One is similar to that of VQA. The other, denoted as \short$_B^*$ in Table~\ref{tab:detailed_result}(d), is first fine-tuned on unbalanced ``all-split'' 
for $5$ epochs, 
and then fine-tuned on the ``balanced-split'' for $2$ epochs, as suggested in~\cite{chen2019meta}.
%




\secvspace
\subsubsection{Natural Language Visual Reasoning for Real (NLVR2)}
\label{sec:NLVR2}~\cite{suhr2018corpus} takes a pair of images and a natural language, and the goal is to determine whether the natural language statement is true about the image pair. 
When fine-tuning on the NLVR2 task, we first construct two input sequences, each containing the concatenation of the given sentence (the natural language description) and one image, and then two $\mathtt{[CLS]}$ outputs from \short{} are concatenated as the joint input for a binary classifier, implemented by an MLP\footnote{This is not necessarily  the best fine-tuning choice for NLVR2, please refer to the \textit{Pair-biattn} finetuning in UNITER~\cite{chen2019uniter} for a better choice, which introduces a multi-head attention layer to look back the concatenated text-image sequences.}.



\section{Experimental Results \& Analysis}
\vspace{-2mm}
\subsection{Performance Comparison with SoTA}
To account for parameter efficiency, we compare \short{} against three types of SoTA's: 
$(\RN{1})$ SoTA$_{S}$ indicates the best performance achieved by small models prior to the Transformer-based VLP models.
$(\RN{2})$ SoTA$_{B}$ indicates the best performance achieved by VLP models of similar size to BERT base. 
$(\RN{3})$ SoTA$_{L}$ indicates the best performance yielded by models that have a similar size to BERT large. To the best of our knowledge, UNITER~\cite{chen2019uniter} is the only model of BERT large size.

Table~\ref{tab:overall_result} summarizes the overall results on all tasks\footnote{All the (single-model) SoTAs are from the published results.}. For all the tables in this paper, \textcolor{blue}{\textbf{Blue}} indicates the best result for a task, and gray background indicates results produced by \short{}. As shown in the table, our base model outperforms previous large models on most tasks, often by a significantly large margin. It demonstrates that the proposed \short{} is highly parameter-efficient, partially because the use of object tags as anchor points significantly eases the learning of semantic alignments between images and texts.
Note that \short{} is pre-trained on $6.5$ million pairs, which is less than $9.6$ million pairs used for UNITER pre-training and $9.18$ million pairs for LXMERT.



\begin{table*}[t!]
\begin{center}
\caption{
Overall results on six tasks. $\Delta$ indicates the improvement over SoTA. SoTA with subscript S, B, L indicates performance achieved by small models, VLP of similar size to BERT base and large model, respectively. Most results are from~\cite{chen2019uniter}, except that image captioning results are from\cite{huang2019attention,zhou2019unified}, NoCaps results are from~\cite{agrawal2019nocaps}, VQA results are from~\cite{tan2019lxmert}.}
\scriptsize
\label{tab:overall_result}
\begin{tabular}{cccc|ccc|cccc|cc|c|c}
\toprule
\multirow{2}{*}{Task} & \multicolumn{3}{c|}{Image Retrieval} & \multicolumn{3}{c|}{Text Retrieval} & \multicolumn{4}{c|}{Image Captioning} & \multicolumn{2}{c|}{NoCaps} & VQA & NLVR2 \\ 
& R@1 & R@5 & R@10 & R@1 & R@5 & R@10 & B@4 & M & C & S & C & S & test-std & test-P\\ \midrule
SoTA$_{S}$ & $39.2$ & $68.0$ & $81.3$ & $56.6$ & $84.5$ & $92.0$ & $38.9$ & $29.2$ & $129.8$ & $22.4$ & $61.5$ & $9.2$ & $70.90$ & $53.50$ \\
SoTA$_{B}$ & $48.4$ & $76.7$ & $85.9$ & $63.3$ & $87.0$ & $93.1$ & $39.5$ & $29.3$ & $129.3$ & $23.2$ & $73.1$ & $11.2$ & $72.54$ & $78.87$\\
SoTA$_{L}$ & $51.7$ & $78.4$ & $86.9$ & $66.6$ & $89.4$ & $94.3$ & $-$ & $-$ & $-$ & $-$ & $-$ & $-$ & $73.40$ & $79.50$ \\
\hline
\rowcolor{Graylight}
\shortb{} & $\bf 54.0$ & $\bf 80.8$ & $\bf 88.5$ & $\bf 70.0$ & $\bf 91.1$ & $\bf 95.5$ & $\bf 40.5$ & $\bf 29.7$ & $\bf 137.6$ & $22.8$ & $\bf 78.8$ & \textcolor{blue}{$\bf 11.7$} & $\bf 73.44$ & $78.36$ \\ 
\rowcolor{Grayheavy}
\shortl{} & \textcolor{blue}{$\bf 57.5$} & \textcolor{blue}{$\bf 82.8$} & \textcolor{blue}{$\bf 89.8$} & \textcolor{blue}{$\bf 73.5$} & \textcolor{blue}{$\bf 92.2$} & \textcolor{blue}{$\bf 96.0$} & \textcolor{blue}{$\bf 41.7$} & \textcolor{blue}{$\bf 30.6$} & \textcolor{blue}{$\bf 140.0$} & \textcolor{blue}{$\bf 24.5$} & \textcolor{blue}{$\bf 80.9$} & $\bf 11.3$ & \textcolor{blue}{$\bf 73.82$} & \textcolor{blue}{$\bf 80.37$} \\
\hline
$\Delta$ & $\bf 5.8\uparrow$ & $\bf 4.4\uparrow$ & $\bf 2.9\uparrow$ & $\bf 6.9\uparrow$ & $\bf 2.8\uparrow$ & $\bf 1.7\uparrow$ & $\bf 2.2\uparrow$ & $\bf 1.3\uparrow$ & $\bf 10.7\uparrow$ & $\bf 1.3\uparrow$ & $\bf 7.8\uparrow$ & $\bf 0.5\uparrow$ & $\bf 0.42\uparrow$ & $\bf 0.87\uparrow$ \\ 
\bottomrule
\vspace{-10mm}
\end{tabular}
\end{center}
\end{table*}

\begin{table*}[t!]
\begin{center}
\scriptsize
\caption{Detailed results on V+L tasks.}
\label{tab:detailed_result}
\begin{tabular}{cc}
\multicolumn{2}{c}{\includegraphics[width=0.98\columnwidth]{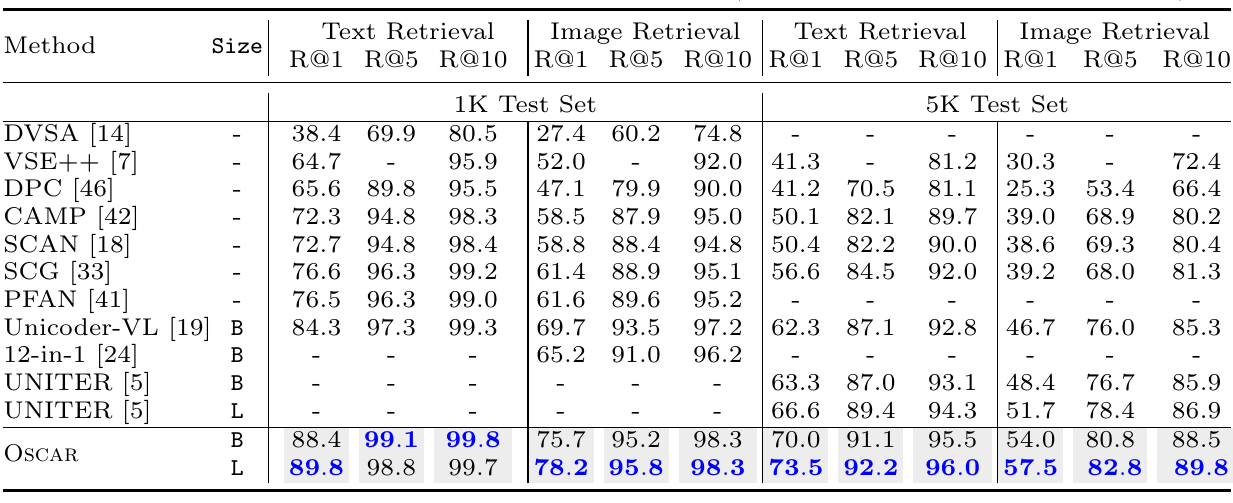}} \\
\multicolumn{2}{c}{(a) Image-text retrieval} \vspace{1mm} \\
\multicolumn{2}{c}{\includegraphics[width=0.94\columnwidth]{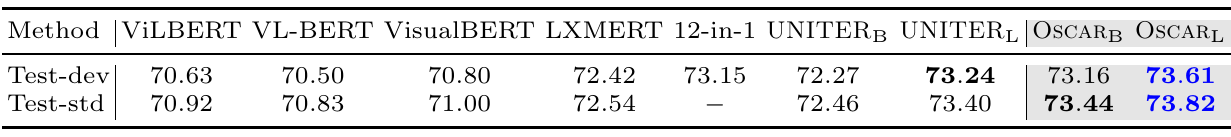}} \\
\multicolumn{2}{c}{(b) VQA} \vspace{1mm}  \\
\multicolumn{2}{c}{\includegraphics[width=0.80\columnwidth]{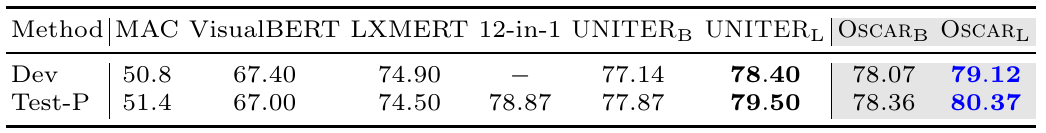}} \\
\multicolumn{2}{c}{(c) NLVR2} \vspace{1mm}  \\
\includegraphics[width=0.30\columnwidth]{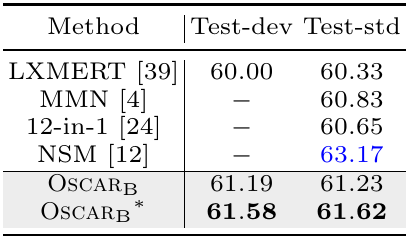}
& 
\includegraphics[width=0.65\columnwidth]{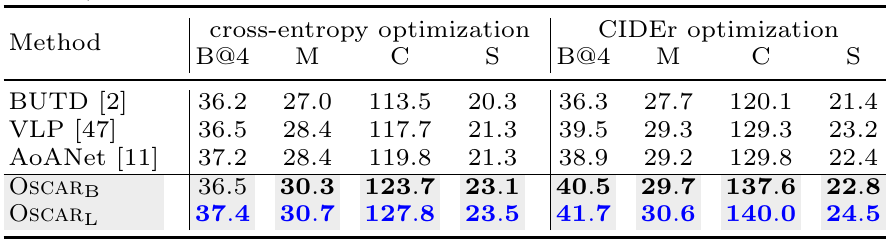} \\
(d) GQA & (e) Image captioning on COCO \vspace{1mm} \\
\multicolumn{2}{c}{\includegraphics[width=0.94\columnwidth]{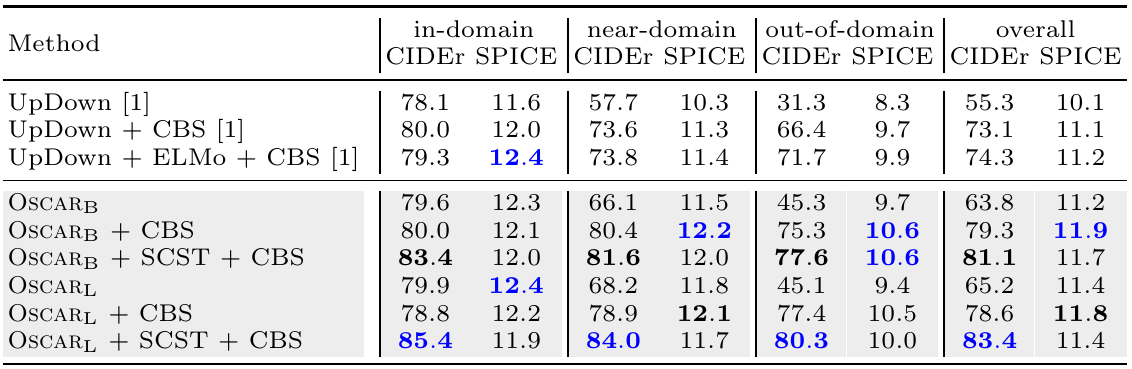}}\\
\multicolumn{2}{c}{(f) Evaluation on NoCaps Val. Models are trained on COCO only without pre-training.}
\vspace{-5mm}
\end{tabular}
\end{center}
\end{table*}

We report the detailed comparison on each task in Table~\ref{tab:detailed_result}. 
$(\RN{1})$ VLP methods dominate empirical performance across many V+L tasks, compared with small models. \short{} outperforms all existing VLP methods on all seven tasks, and achieves new SoTA on six of them. On GQA, neural state machine (NSM)~\cite{hudson2019learning} relies on a strong structural prior, which can also be incorporated into \short{} for improvement in the future.
$(\RN{2})$ 12-in-1 is a recently proposed multi-task learning model~\cite{lu201912} for V+L, implemented on BERT base. We see that \shortb{} outperforms 12-in-1 on almost all the tasks, except on Test-P of NLVR2. Given that our method is based on single task fine-tuning, the result demonstrates the effectiveness of our proposed pre-training scheme.
$(\RN{3})$ overall, \short{} is the best performer on both understanding and generation tasks. 
On the captioning task, we further fine-tune \short{} with self-critical sequence training (SCST)~\cite{rennie2017self} to improve sequence-level learning. The only comparable VLP method for captioning is~\cite{zhou2019unified}. 
The results in Table~\ref{tab:detailed_result} (e) show that \short{} yields a much better performance, \eg improving BLEU@4 and CIDEr by more than 2 and 10 points, respectively.
$(\RN{4})$ The NoCaps guideline requires to only use the COCO captioning training set. Hence, we initialize with BERT, and train \short{} on the COCO training set. Constrained beam search (CBS) is used. The results in Table~\ref{tab:detailed_result} (f) show that the variants of \short{} consistently outperform the previous SoTA method UpDown~\cite{agrawal2019nocaps}. The gap is much larger on the near-domain or out-of-domain cases, demonstrating the strong generalization ability of \short{}.

\subsection{Qualitative Studies}

\begin{figure*}[t!]
\centering
\begin{tabular}{c c}
	\hspace{-2mm}
	\includegraphics[height=4.0cm]{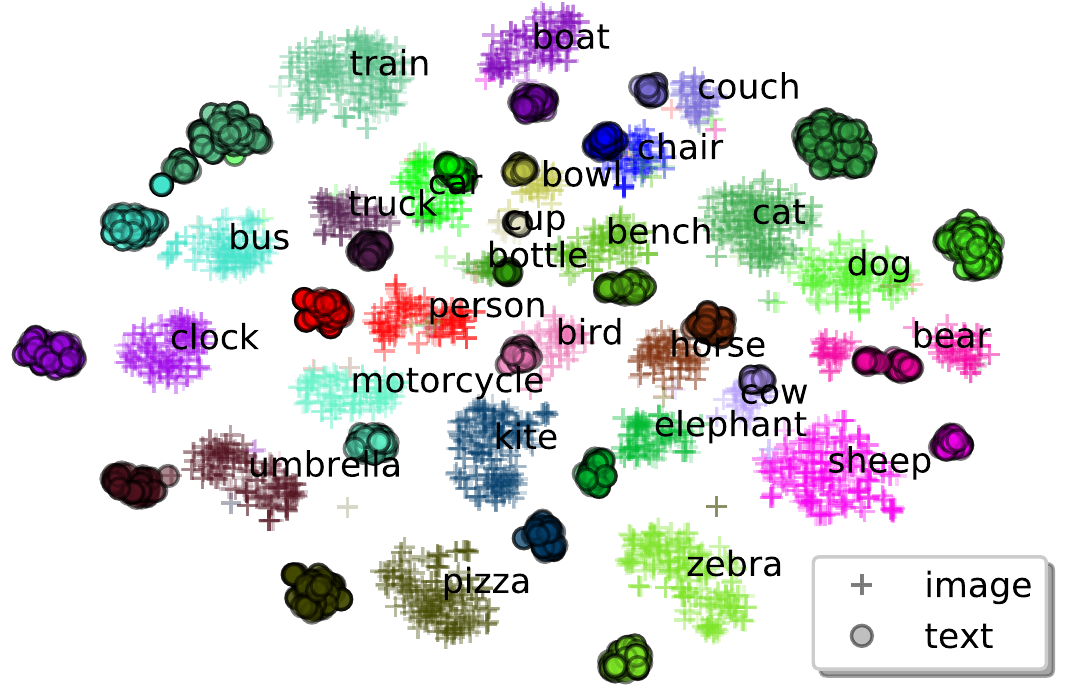} & 
	 \hspace{-0mm}
	\includegraphics[height=4.0cm]{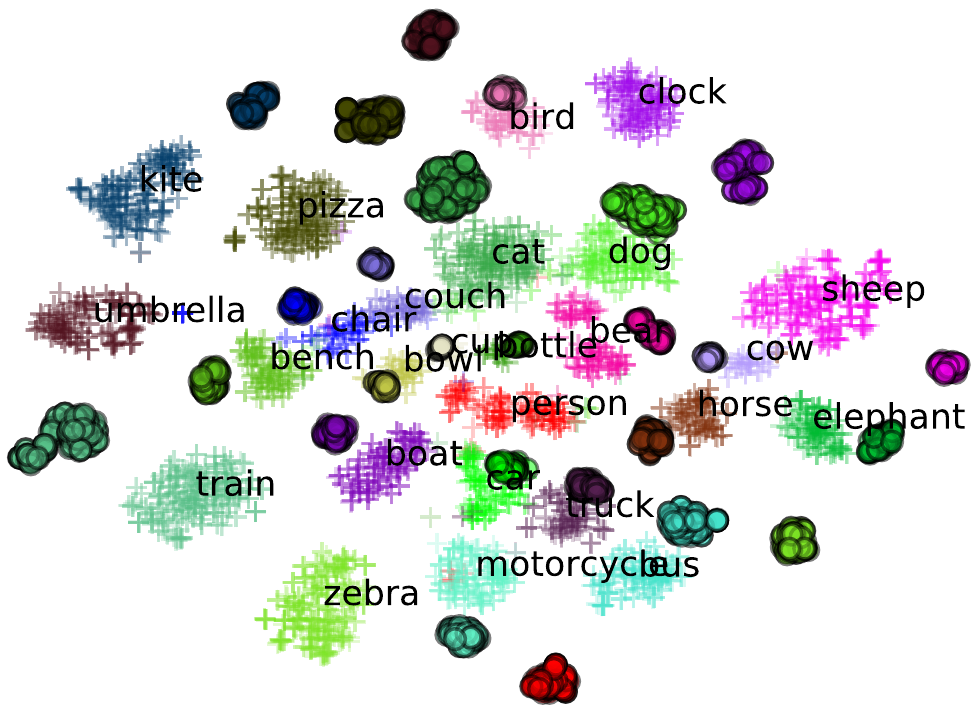}
	\\
	(a) \short{} \vspace{2mm} & 
	\hspace{2mm}  
	(b) Baseline (No tags) \hspace{-0mm}
\end{tabular}
\vspace{-2mm}
\caption{2D visualization using $t$-SNE. The points from the same object class share the same color. Please refer Appendix for full visualization.}
\label{fig:tnse}
\end{figure*}

We visualize the learned semantic feature space of image-text pairs of the COCO test set on a 2D map using $t$-SNE~\cite{maaten2008visualizing}. For each image region and word token, we pass it through the model, and use its last-layer output as features. Pre-trained models with and without object tags are compared. The results in Fig~\ref{fig:tnse} reveal some interesting findings.
$(\RN{1})$ {\it Intra-class.} With the aid of object tags, the distance of the same object between two modalities is substantially reduced. For example, the visual and textual representations for $\mathtt{person}$ (or $\mathtt{zebra}$) in \short{} is much closer than that in the baseline method.
$(\RN{2})$ {\it Inter-class.}
Object classes of related semantics are getting closer (but still distinguishable) after adding tags, while there are some mixtures in the baseline, such as animal ($\mathtt{person}$,  $\mathtt{zebra}$,  $\mathtt{sheep}$,  $\mathtt{bird}$), furniture ($\mathtt{chair}$, $\mathtt{couch}$, $\mathtt{bench}$), and transportation ($\mathtt{bus}$, $\mathtt{train}$, $\mathtt{truck}$, $\mathtt{motorcycle}$, $\mathtt{car}$). This verifies the importance of object tags in alignment learning: it plays the role of anchor points in linking and regularizing the cross-modal feature learning.

\begin{figure}[t!]
\centering
\includegraphics[width=0.9\columnwidth]{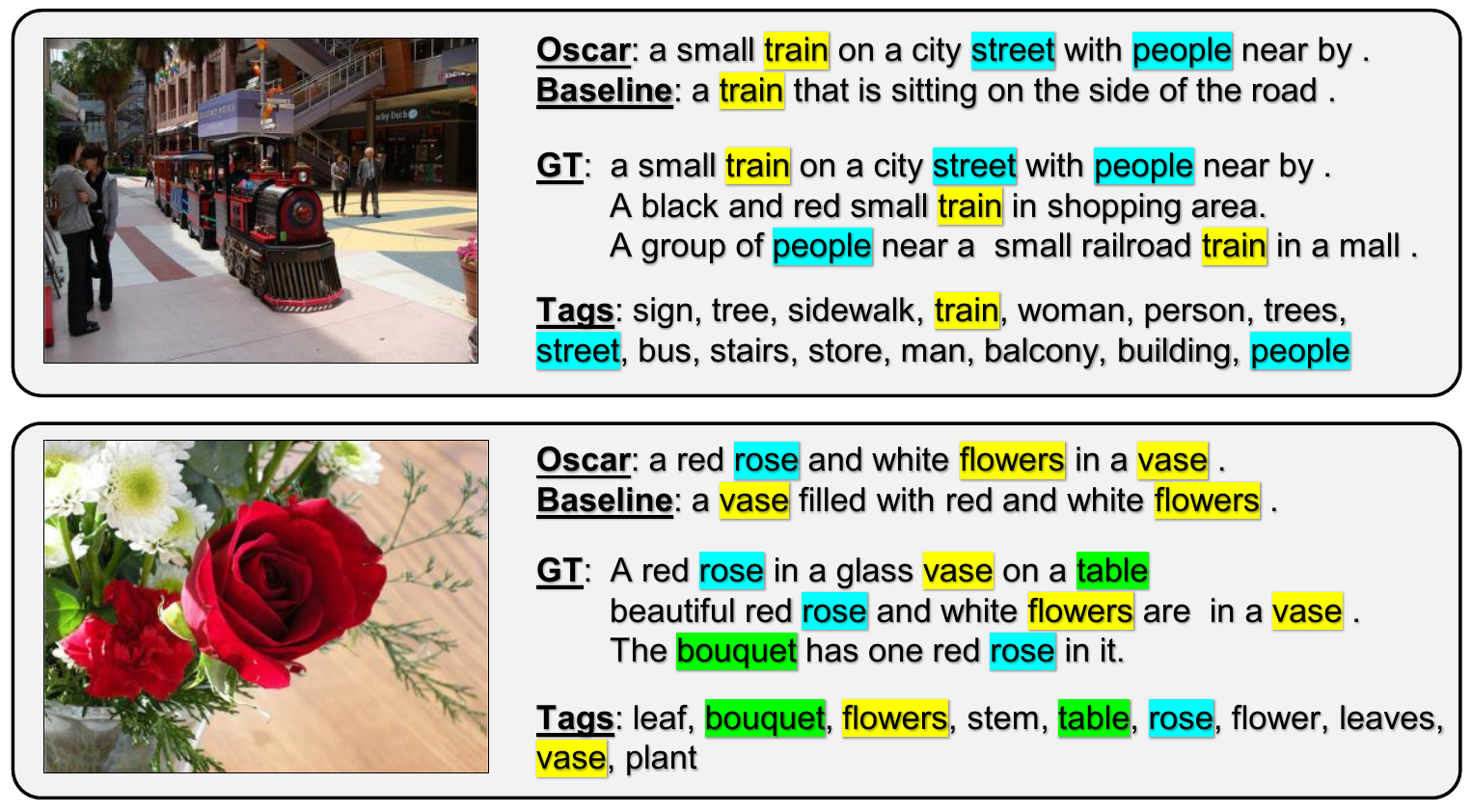}
\caption{Examples of image captioning. Objects are colored, based on their appearance against the groud-truth (GT): \colorbox{yellow}{all},  \colorbox{skyblue}{\short{} \& tags}, \colorbox{lightgreen}{tags only}. }
\label{fig:imgcap_examples}
\vspace{-4mm}
\end{figure}

We compare generated captions of different models in Fig.~\ref{fig:imgcap_examples}. The baseline method is VLP without object tags. We see that \short{} generates more detailed descriptions of images than the baseline, due to the use of the accurate and diverse object tags detected by Faster R-CNN. They are the anchor points in the word embedding space, guiding the text generation process.

\subsection{Ablation Analysis}
We perform ablation experiments over a number of design choices of \short{} in both pre-training and fine-tuning to better understand their relative importance to four representative downstream tasks. All the ablation experiments are conducted on the base model.

\begin{figure*}[t!]
\centering 
\subfigure[VQA] { 
\label{fig:learning_typeA} 
\includegraphics[width=0.31\columnwidth]{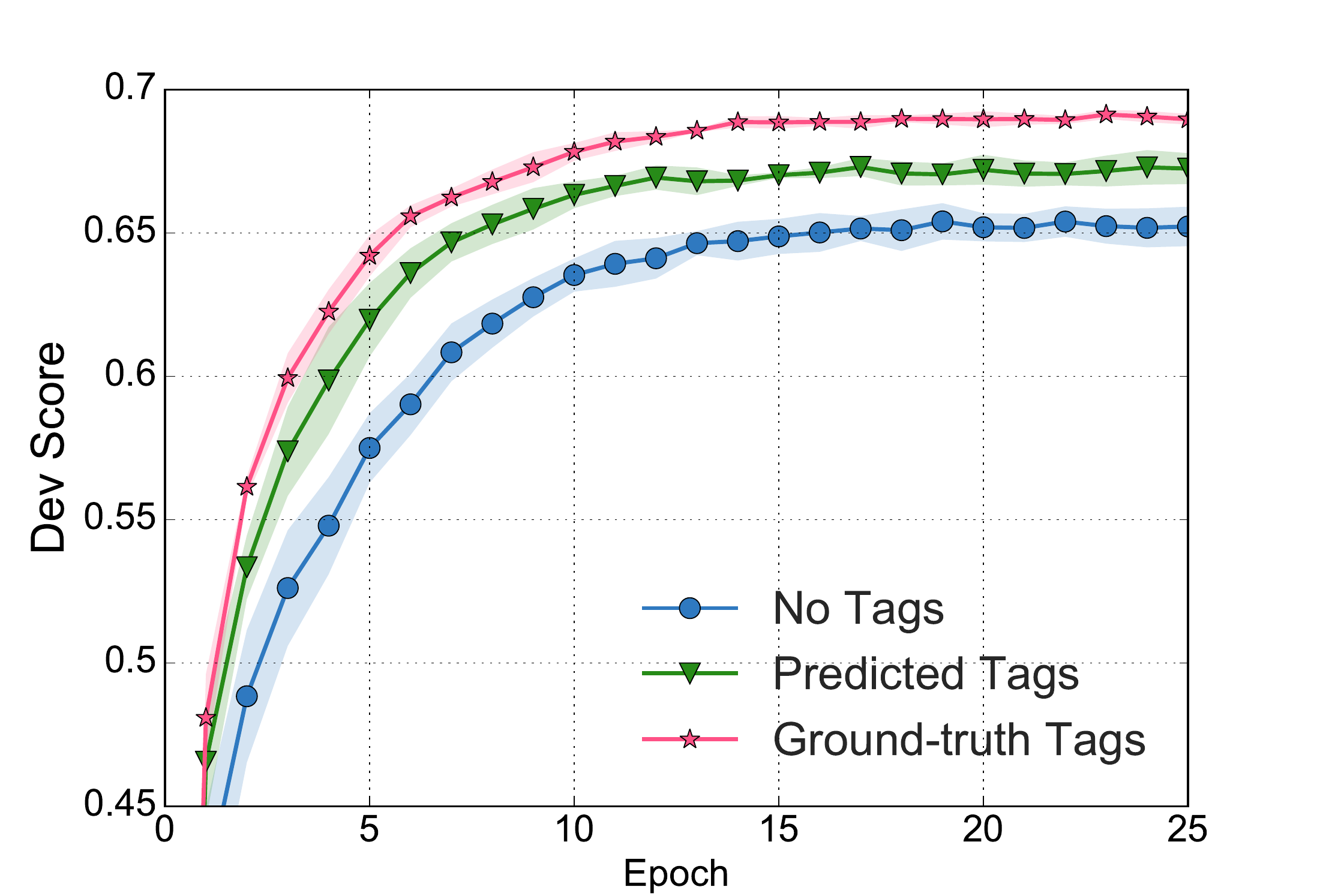}}
\subfigure[Image Retrieval R@1] { 
\label{fig:learning_typeB} 
\includegraphics[width=0.31\columnwidth]{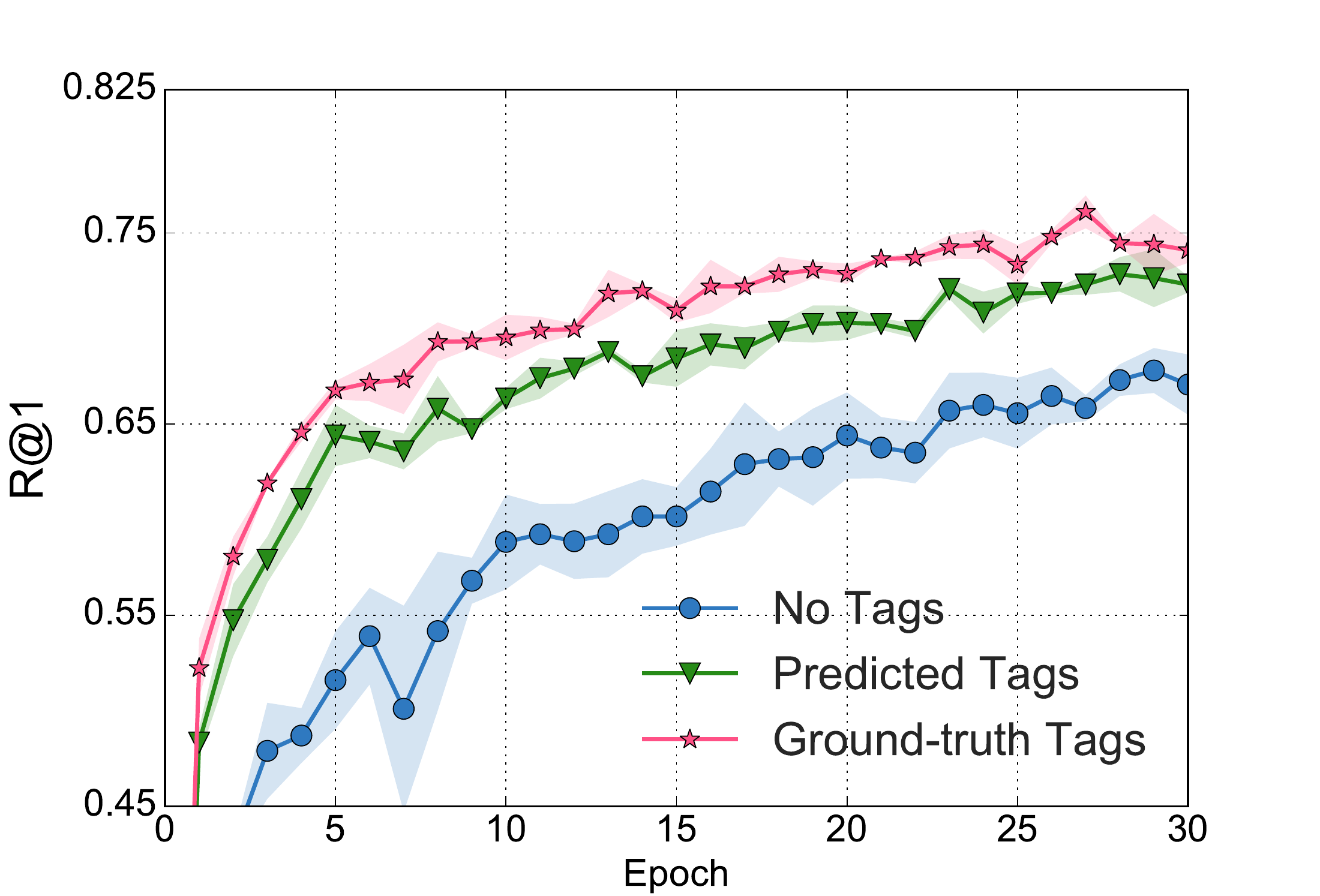}}
\subfigure[Image Captioning] { 
\label{fig:learning_typeB} 
\includegraphics[width=0.31\columnwidth]{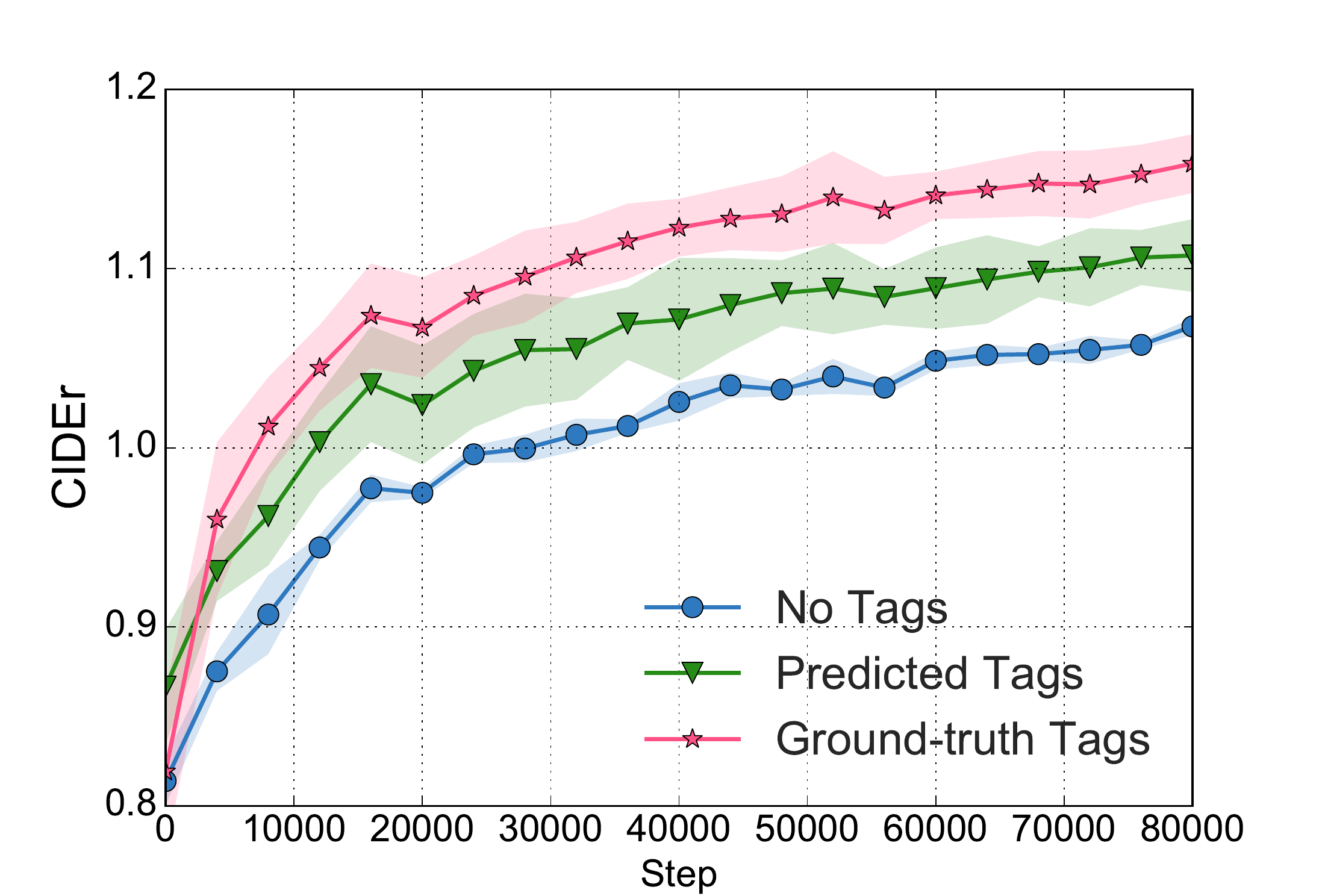}}
\vspace{-3mm}
\caption{The learning curves of fine-tuning downstream tasks with different object tags. Each curve is with 3 runs.} 
\label{fig:ob_abl} 
\vspace{-2mm}
\end{figure*}

\secvspace
\subsubsection{The Effect of Object Tags}
To study the effect of object tags, we experiment three different settings: 
$(\RN{1})$ \textit{Baseline (No Tags)}: this reduces the models to their previous VLP counterparts, where no tag information is exploited.
$(\RN{2})$ \textit{Predicted Tags}: we use an off-the-shelf object detector (trained on COCO dataset) to predict object tags.
$(\RN{3})$ \textit{Ground-truth Tags}: The ground-truth tags from COCO dataset are utilized to serve as a performance ``upper bound'' for our method. The experiments are conducted with the same BERT base model on three representative tasks, including VQA, image retrieval, and image captioning. 
As shown in Fig.~\ref{fig:ob_abl}, the learning curves for fine-tuning with object tags converges significantly faster and better than the VLP method without tags on all tasks. On the VQA and retrieval tasks, training using tags only takes half of the training time to achieve the final performance of the baseline, showing that \short{} is a more practical and efficient scheme for VLP. With more accurate object detectors developed in the future, \short{} can achieve even better performance, closing the gap demonstrated by using the ground-truth tags.

\begin{wrapfigure}{R}{5.00cm}
\centering
	\vspace{-8mm}
	\captionof{table}{\small Retrieval results on the COCO 1K test set, with different types of attention interactions.}
	\begin{tabular}{c}	
	\hspace{-4mm}
		\includegraphics[width=4.40cm]{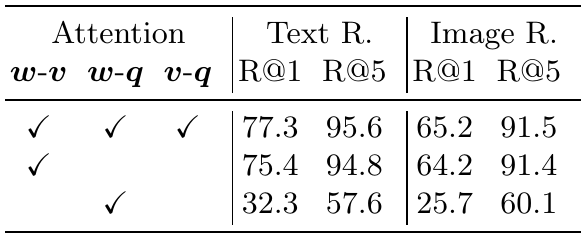} \\
	\end{tabular}
	\vspace{-2mm}
	\label{tab:ab_attention}
\vspace{-6mm}
\end{wrapfigure}
\subsubsection{Attention Interaction}
To further understand the interaction among the text, object tags and object regions, we conduct fine-tuning experiments by varying the attention masks for image-text retrieval. The default setting uses full attentions across all modalities. We then enable certain part of the attention masks. All models are initialized from BERT base without pre-training. 
Table~\ref{tab:ab_attention} reports the performance on the COCO $1$K test set. By comparing the results of using full attention and partial attention $\wv$-$\vv$, we see that it is beneficial to add object tags. Moreover, region features are more informative than object tags ($\wv$-$\vv$, vs.  $\vv$-$\qv$) in representing an image. This suggests that tags yield minor improvement when used as features; a more promising way is to use them as anchor points, as done in \short{}.

\secvspace
\subsubsection{Object Tags in Pre-training}
To study the impact of different object tag sets in pre-trained models, we pre-train two variants: \short{}$^{\text{VG}}$ and \short{}$^{\text{OI}}$ utilizes object tags produced by the object detector trained on the visual genome (VG) dataset~\cite{krishna2017visual} and the open images (OI) dataset~\cite{kuznetsova2018open}, respectively. In this ablation, all the models are pre-trained for 589k steps. The results are shown in Table~\ref{tab:ab_schema}, where \textsc{Baseline (No Tags)} is also listed for comparison. It is clear that the \short{} scheme of using object tags as anchor points improves the baseline, regardless of which set of object tags is used. VG tags performs slightly better than OI. We hypothesize that the object detector trained on VG has a more diverse set of objects, although the object detector trained on OI has a higher precision. 


\begin{table*}[t!]
\begin{center}
\caption{
Results with various pre-training schemes. 
}
\scriptsize
\label{tab:ab_schema}
\begin{tabular}{l@{\hspace{1.5mm}}|c@{\hspace{1.5mm}}|c@{\hspace{1.5mm}}c@{\hspace{1.5mm}}c@{\hspace{1.5mm}}|c@{\hspace{1.5mm}}c@{\hspace{1.5mm}}c@{\hspace{1.5mm}}|c@{\hspace{1.5mm}}c@{\hspace{1.5mm}}c@{\hspace{1.5mm}}c}
\toprule
\multirow{2}{*}{Pre-train} & VQA & \multicolumn{3}{c|}{Text Retrieval} & \multicolumn{3}{c|}{Image Retrieval} & \multicolumn{4}{c}{Image Captioning} \\ 
& dev & R@1 & R@5 & R@10 & R@1 & R@5 & R@10 & B@4 & M & C & S\\ \midrule
\textsc{Baseline (No Tags)} & $70.93$ & $84.4$ & $98.1$ & $99.5$ & $73.1$ & $94.5$ & $97.9$ & $34.5$ & $29.1$ & $115.6$ & $21.9$\\
\short$^{\text{VG}}$ & $71.70$ & $88.4$ & $99.1$ & $99.8$ & $75.7$ & $95.2$ & $98.3$ & $36.4$ & $30.3$ & $123.4$ & $23.0$\\ 
\short$^{\text{OI}}$ & $71.15$ & $85.9$ & $97.9$ & $99.5$ & $72.9$ & $94.3$ & $97.6$ & $35.3$ & $29.6$ & $119.5$ & $22.6$\\ 
\bottomrule
\end{tabular}
\end{center}
\vspace{-5mm}
\end{table*}

\section{Related Work}

\subsubsection{Vision-Language Pre-training} There is a growing interest in pre-training generic models to solve a variety of V+L problems, such as visual question-answering (VQA), image-text retrieval and image captioning~\etc~
The existing methods~\cite{sun2019videobert,tan2019lxmert,lu2019vilbert,chen2019uniter,zhou2019unified,su2019vl,li2019unicoder,hao2020prevalent} employ BERT-like objectives~\cite{devlin2019bert} to learn cross-modal representations from a concatenated-sequence of visual region features and language token embeddings. They heavily rely on the self-attention mechanism of Transformers to learn joint representations that are appropriately contextualized in both modalities. For example, early efforts such as \cite{lu2019vilbert,tan2019lxmert} propose a two-stream and three-stream Transformer-based framework with co-attention to fuse the two modalities, respectively. Chen \ea \cite{chen2019uniter} conduct comprehensive studies on the effects of different pre-training objectives for the learned generic representations. Zhou \ea \cite{zhou2019unified} propose the first unified model to deal with both understanding and generation tasks, using only VQA and image captioning as the downstream tasks. In this paper, the \short{} models have been applied to a wider range of downstream tasks, including both understanding and generation tasks, and have achieved new SoTA in most of them.  
Compared to existing VLP methods, the most salient difference of the proposed \short{} is the use of object tags for aligning elements in two modalities. It alleviates the challenge of VLP models having to figure out the cross-modal semantic alignment from scratch, and thus improves the learning efficiency. In fact, our base model already outperforms the existing large VLP models on most V+L tasks.

\secvspace
\subsubsection{Object Tags} Anderson \ea ~\cite{anderson2018bottom} introduce the bottom-up mechanism to represent an image as a set of visual regions via Faster R-CNN~\cite{ren2015faster}, each with an associated feature vector. It enables attention to be computed at the object level, and has quickly become the de facto standard for fine-grained image understanding tasks. 
In this paper, we propose to use object tags to align the object-region features in~\cite{anderson2018bottom} in the pre-trained linguistic semantic space. The idea of utilizing object tags has been explored for image understanding~\cite{wu2016value,you2016image,zhou2019unified}. Based on grid-wise region features of CNNs, Wu \ea~\cite{wu2016value} employ the predicted object tags only as the input to LSTM for image captioning, while You \ea~\cite{you2016image} consider both tags and region features. Based on salient regions proposed by object detectors, Zhou \ea~\cite{zhou2019unified} concatenate the object prediction probability vector with region features as the visual input for VLP. Unfortunately, the tags in these works are not simultaneously associated with both object regions and word embeddings of text, resulting in a lack of grounding. Our construction of object tags with their corresponding region features \& word embeddings yields more complete and informative representations for objects, particularly when the linguistic entity embeddings are pre-trained, as described next. 

\secvspace
\subsubsection{Multimodal Embeddings} It has been shown that V+L tasks can benefit from a shared embedding space to align the inter-modal correspondences between images and text. Early attempts from Socher \ea~\cite{socher2010connecting} project words and image regions into a common space using kernelized canonical correlation analysis, and achieve good results for annotation and segmentation. Similar ideas are employed for image captioning~\cite{karpathy2015deep} and text-based image retrieval~\cite{ren2016joint}. 
In particular, the seminal work DeViSE~\cite{frome2013devise} proposes to identify visual objects using semantic information gleaned from un-annotated text. This semantic information is exploited to make predictions of image labels that are not observed during training, and improves zero-shot predictions dramatically across thousands of novel labels that have never been seen by the vision model. The idea has been extended  in~\cite{socher2013zero,kiros2014unifying,norouzi2013zero}, showing that leveraging pre-trained linguistic knowledge is highly effective for aligning semantics and improving sample efficiency in cross-modal transfer learning. Inspired by this line of research, we revisit the idea and propose to leverage the rich semantics from the learned word embeddings in the era of neural language model pre-training. Indeed, our results on novel object captioning demonstrate that \short{} helps improve the generalizability of the pre-trained models.
\section{Conclusion}
In this paper, we have presented a new pre-training method \short{}, which uses object tags as anchor points to align the image and language modalities in a shared semantic space. We validate the schema by pre-training \short{} models on a public corpus with 6.5 million text-image pairs. The pre-trained models archive new state-of-the-arts on six established V+L understanding and generation tasks.


%
%
\bibliographystyle{splncs04}
\bibliography{egbib}

\appendix
\newpage
\section{Fine-tuning Settings}

\vspace{-2mm}
\subsubsection{Image-Text Retrieval}
We adopt the widely used Karpathy split~\cite{karpathy2015deep} on the COCO caption dataset~\cite{lin2014microsoft} to conduct our experiments. Specifically, the dataset consists of $113,287$ images for training, $5,000$ images for validation, and $5,000$ images for testing. Each image is associated with $5$ human-generated captions. For the \shortb{} model, we fine-tune with a batch size of $256$ for $40$ epochs. The initial learning rate is set to $2e^{-5}$ and linearly decreases. For the \shortl{} model, we fine-tune with a batch size of $128$ for $40$ epochs. The initial learning rate is set to $1e^{-5}$ and linearly decreases. We use the validation set for parameter tuning. 
We compare with several existing methods, including
DVSA~\cite{karpathy2015deep},
VSE++~\cite{faghri2017vse++}, 
DPC~\cite{zheng2017dual}, 
CAMP~\cite{wang2019camp}, 
SCAN~\cite{lee2018stacked}, 
SCG~\cite{shi2019knowledge}, 
PFAN~\cite{wang2019position}, 
Unicoder-VL~\cite{li2019unicoder},
12-in-1~\cite{lu201912},
UNITER~\cite{chen2019uniter}.

\suppvspace
\subsubsection{Image Captioning}
Though the training objective (\ie seq2seq) for image captioning is different from that used in pre-training (\ie bidirectional attention-based mask token loss), we directly fine-tune \short{} for image captioning on COCO without additional pre-training on Conceptual Captions~\cite{sharma2018conceptual}. This is to validate the generalization ability of the \short{} models for generation tasks. We use the same Karpathy split~\cite{karpathy2015deep}. During training, we randomly select $15\%$ of caption tokens with a maximum of $3$ tokens per caption to be masked out. For the \shortb{} model, we fine-tune with cross-entropy loss for $40$ epochs with a batch size of $256$ and an initial learning rate of $3e^{-5}$ and then with CIDEr optimization~\cite{rennie2017self} for $5$ epochs with a batch size of $64$ and initial learning rate of $1e^{-6}$. For the \shortl{} model,  we fine-tune for $30$ epochs with a batch size of $128$ and an initial learning rate of $1e^{-5}$ and then with CIDEr optimization for another $3$ epochs with a batch size of $48$ and learning rate of \{$1e^{-6}$, $5e^{-7}$\}. 
We compare with several existing methods, including
BUTD~\cite{anderson2018bottom},
VLP~\cite{zhou2019unified},
AoANet~\cite{huang2019attention}.

\suppvspace
\subsubsection{NoCaps}
Since NoCaps images are collected from Open Images. We train an object detector using the Open Images training set and applied it to generate the tags. We conduct experiments from BERT model directly without pre-training as required by the task guidelines. For the \shortb{} model, we train $40$ epoch with a batch size of $256$ and learning rate $3e^{-5}$; further we perform CIDEr optimization with learning rate $1e^{-6}$ and batch size $64$ for $5$ epochs. During inference, we use constrained beam search for decoding. We compare \short{} with UpDown~\cite{agrawal2019nocaps} on this task.

\suppvspace
\subsubsection{VQA}
For VQA training, we random sample a set of 2k images from the MS COCO validation set as our validation set, the rest of images in the training and validation are used in the VQA finetuning. For the \shortb{} model, we fine-tune for $25$ epochs with a learning rate of $5e^{-5}$ and a batch size of $128$. For the \shortl{} model, we fine-tune for $25$ epochs with with a learning rate of $3e^{-5}$ and a batch size of $96$. 

\suppvspace
\subsubsection{GQA}
The fine-tuning procedure of GQA is similar to that of VQA. For the \shortb{} model, we fine-tune for $5$ epochs with a learning rate of $5e^{-5}$ and a batch size of $128$. We compare with four existing methods, including LXMERT~\cite{tan2019lxmert},
MMN~\cite{chen2019meta},
12-in-1~\cite{lu201912},
NSM~\cite{hudson2019learning}.

\suppvspace
\subsubsection{NLVR2}
For the \shortb{} model, we fine-tune for $20$ epochs with learning rate \{$2e^{-5}$, $3e^{-5}$, $5e^{-5}$\} and a batch size of $72$. For the \shortl{} model, we fine-tune for $20$ epochs with learning rate of \{$2e^{-5}$, $3e^{-5}$\} and a batch size of $48$.

\section{Pre-training Corpus}
Table~\ref{tab:pretrain_corpus} shows the statistics of image and text of the corpus.
\begin{table}[h]
\begin{center}
\vspace{-5mm}
\caption{Statistics of the pre-training corpus.}
\scriptsize
\label{tab:pretrain_corpus}
\begin{tabular}{c|c|c|c|c|c|c|c|c}
\toprule
\multirow{2}{*}{Source} & COCO & CC & SBU & Flicker30k & VQA & GQA & VG-QA & Total \\
 & (train) & (all) & (all) & (train) & (train) & (bal-train) & (train) &  \\ 
\midrule
Image/Text & 112k/560k & 3.0M/3.0M & 840k/840k & 29k/145k & 83k/444k & 79k/1026k & 48k/484k & 4.1M/6.5M  \\
\bottomrule
\end{tabular}
\end{center}
\vspace{-5mm}
\end{table}

\vspace{-3mm}
\section{More Results}
\begin{figure}[t!]
\centering
{\includegraphics[width=0.88\textwidth]{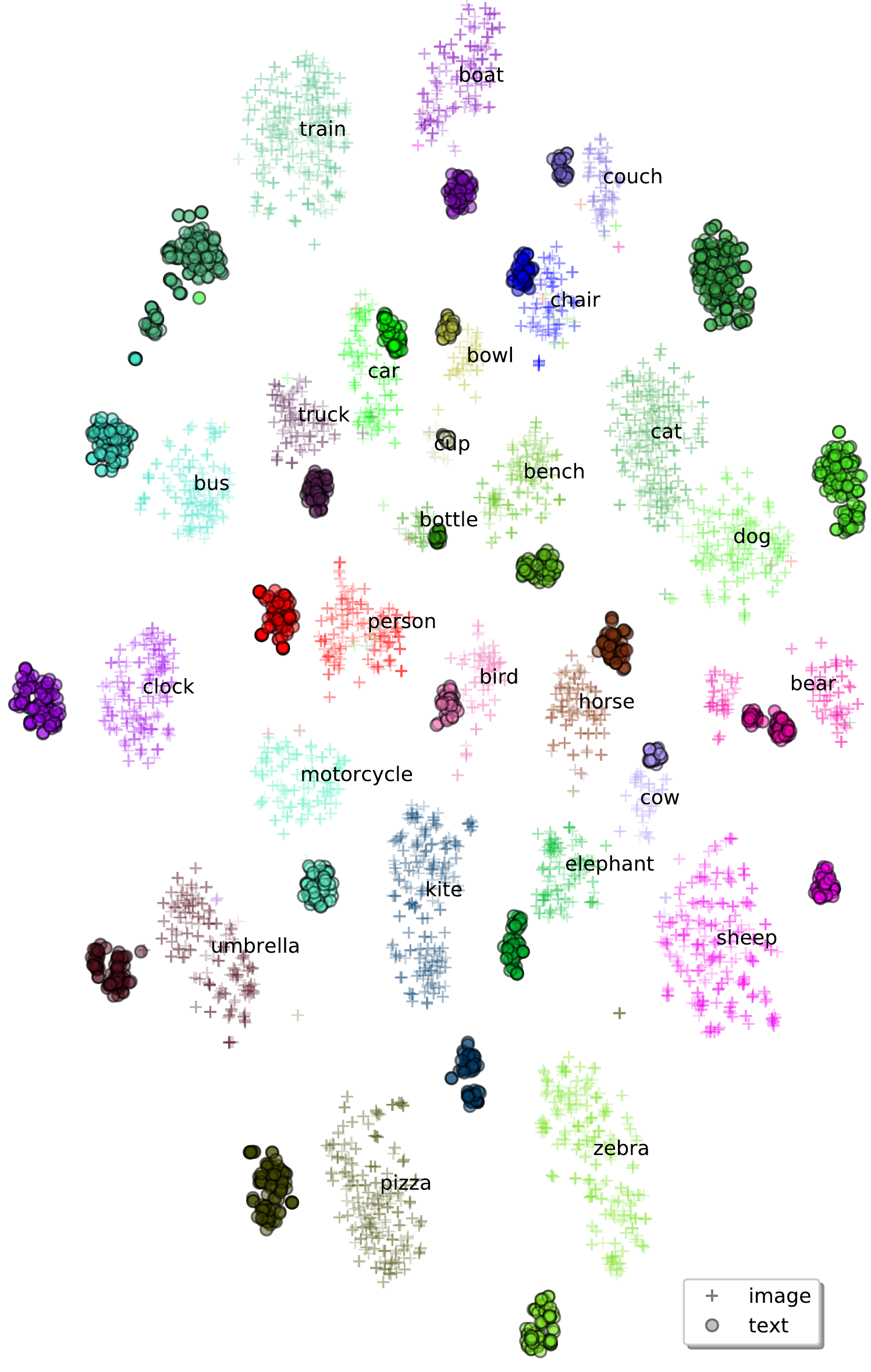}}
 \vspace{-0mm}
\caption{Feature visualization of \short{}. We observe small distances between text and image features of the same object; some of them are perfectly aligned, as demonstrated by the overlapping regions.}
\label{fig:tsne_oscar_big}
 \vspace{-0mm}
 \end{figure}

\begin{figure}[t!]
\centering
{\includegraphics[width=0.82\textwidth]{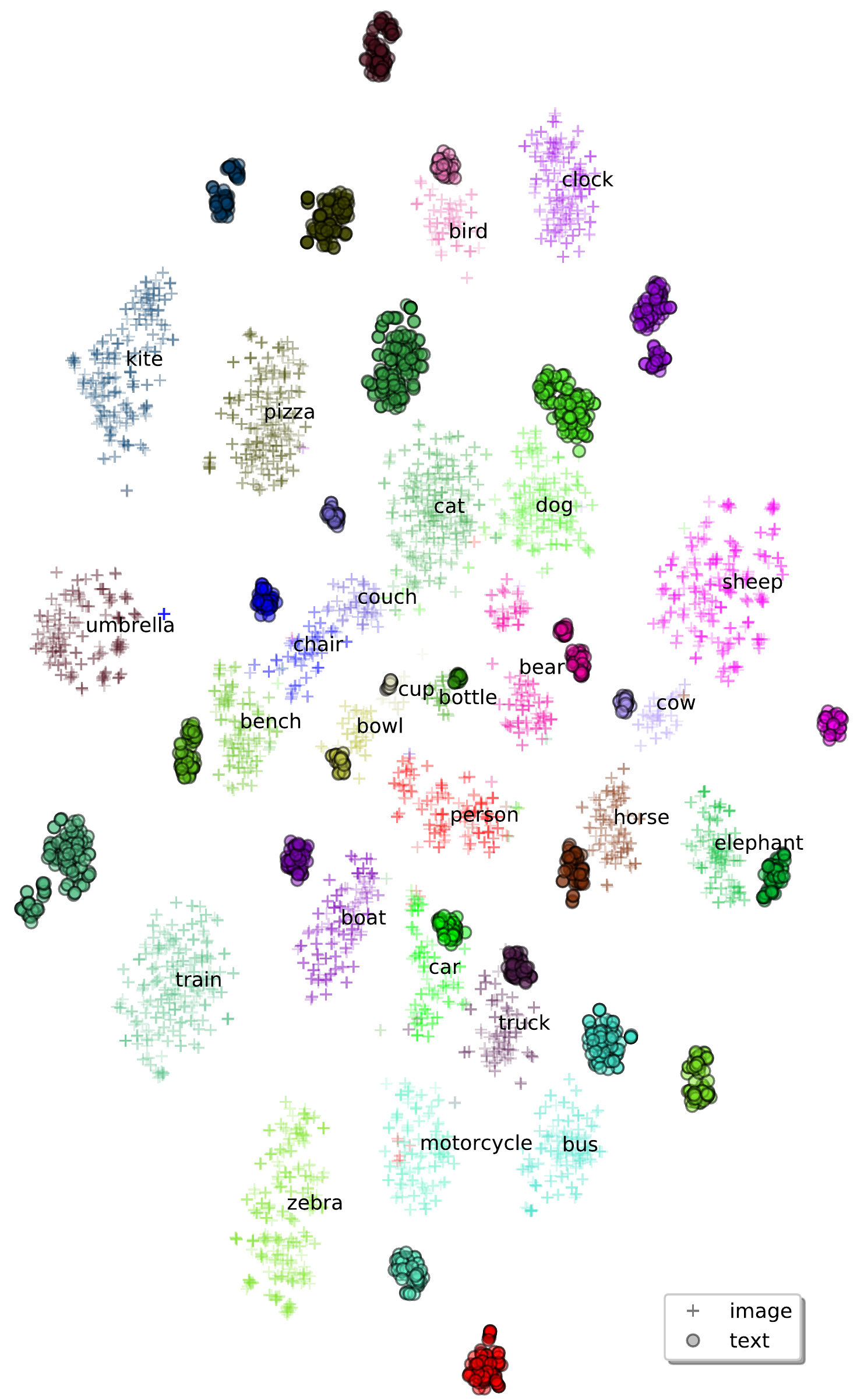}}
 \vspace{-0mm}
\caption{Feature visualization of baseline (no tags). For several object classes, their text and image features are largely separated (\eg person, umbrella, zebra). The distance of image features between some objects is too small (\eg bench, chair, couch).}
\label{fig:tsne_baseline_big}
 \vspace{-0mm}
 \end{figure}

The enlarged $t$-SNE visualization results of \short{} and baseline (no tags) are shown in Fig.~\ref{fig:tsne_oscar_big} and Fig.~\ref{fig:tsne_baseline_big}, respectively.


\section*{Acknowledgement}
We thank Yonatan Bisk, Hannaneh Hajishirzi, Xiaodong Liu, Sachin Mehta, Hamid Palangi and Arun Sacheti, Rowan Zellers for valuable discussions and comments, and the Microsoft Research Technical Support team for managing the GPU clusters.



\end{document}